# Explainable machine learning-based prediction model for diabetic nephropathy

Jing-Mei Yin,4* Yang Li,1* Jun-Tang Xue,1 Guo-Wei Zong,2,3 Zhong-Ze Fang,1,3 and Lang Zou4

1Department of Toxicology and Sanitary Chemistry, School of Public Health, Tianjin Medical University, Tianjin, China

2Department of Mathematics, School of Public Health, Tianjin Medical University, Tianjin, China

3Tianjin Key Laboratory of Environment, Nutrition and Public Health, Tianjin, China

4 School of Mathematics and Computational Science Xiangtan University, Xiangtan, Hunan, China

*Jing-Mei and Yang Li have contributed equally to this work and share first authorship.

Correspondences:

Zhong-Ze Fang, Department of Toxicology and Sanitary Chemistry, School of Public Health, Tianjin Medical University, Tianjin, China; Fax: 83336122; E-mail: fangzhongze@tmu.edu.cn.

Lang Zou, School of Mathematics and Computational Science Xiangtan University, Xiangtan, Hunan, China; E-mail: langzou@xtu.edu.cn.

Abstract:

The aim of this study is to analyze the effect of serum metabolites on diabetic nephropathy (DN) and predict the prevalence of DN through a machine learning approach. The dataset consists of 548 patients from April 2018 to April 2019 in Second Affiliated Hospital of Dalian Medical University (SAHDMU). We select the optimal 38 features through a Least absolute shrinkage and selection operator (LASSO) regression model and a 10-fold cross-validation. We compare four machine learning algorithms, including eXtreme Gradient Boosting (XGB), random forest, decision tree and logistic regression, by AUC-ROC curves, decision curves, calibration curves. We quantify feature importance and interaction effects in the optimal predictive model by Shapley Additive exPlanations (SHAP) method. The XGB model has the best performance to screen for DN with the highest AUC value of 0.966. The XGB model also gains more clinical net benefits than others and the fitting degree is better. In addition, there are significant interactions between serum metabolites and duration of diabetes. We develop a predictive model by XGB algorithm to screen for DN. C2, C5DC, Tyr, Ser, Met, C24, C4DC, and Cys have great contribution in the model, and can possibly be biomarkers for DN.

Keywords: diabetic nephropathy, serum metabolites, machine learning, risk prediction

Key Summary Points

Why carry out this study ?

The prevalence of diabetic nephropathy has been increasing in recent years, but there are few screening methods for it.

What was learned from the study ?

The prediction model based on XGB algorithm shows that C2, C5DC, Tyr, Ser, Met, C24, C4DC, and Cys have high correlation with DN.

Patients with longer diabetes duration and lower C5DC values had a lower risk of disease compared to those with higher C5DC values.

Patients with longer diabetes duration and lower Tyr values had a higher risk of disease compared to those with higher Tyr values.

# Introduction

Diabetes mellitus is an extremely common chronic disease. By 2045, the prevalence of diabetes will rise to 10.9%.[1] Of greater concern to us is that the Western Pacific will have the highest number of adult diabetics in the world.[3] In China, about 20-40% of diabetic patients have combined renal complications, and diabetic nephropathy (DN) has become the leading cause of end-stage chronic kidney disease.[49] Meanwhile, the all-cause mortality rate in patients with DN is nearly 20-40 times higher than that in non-diabetic nephropathy.[2] New screening and treatment methods have important implications for the prevention of diabetic nephropathy in the country.

In recent years, there has been a growing interest in metabolomic measurements to identify pathophysiological mechanisms, new diagnostic and prognostic biomarkers associated with disease development.[4] Among the various serum metabolites that have been extensively studied, amino acids and acylcarnitine have received much attention in recent years. Amino acids are involved in different physiological roles of the body, such as cell signaling, gene expression, nutrient metabolism and endocrine hormone production.[5] There is research evidence that dysregulation of acylcarnitine homeostasis plays a role in the development and progression of various diseases, such as insulin resistance and metabolic syndrome.[6,7]

Since traditional clinical indicators and serum metabolites have a large number of features and are high-dimensional datasets containing both correlated and uncorrelated data, it is not sufficient to analyze such data using traditional statistical methods.[8] In recent years, machine learning methods, such as Least absolute shrinkage and selection operator

(LASSO) regression, Support Vector Machine (SVM), Decision Tree (DT), Random Forest (RF), and Artificial Neural Networks (NNs) have been widely used in healthcare,[9] such as cancer, medicinal chemistry, medical imaging, etc.[10] Investigations have shown that machine learning can help improve the reliability, performance, predictability and accuracy of diagnostic systems for diseases that require it, and can be used to examine important clinical parameters, biological indicators and serum metabolites.[11,12]

The purpose of this paper is to develop and test a prediction model for DN by using machine learning methods and the dataset of Dalian Second People's Hospital, and explain the prediction model to quantify the influence of serum metabolites to DN.

# Material and methods

## *Data*

### Data source

Data for this paper including 1024 participants are obtained from April 2018 to April 2019 in Second Affiliated Hospital of Dalian Medical University (SAHDMU). Demographic parameters, anthropometric, clinical and laboratory parameters, medications and disease conditions are extracted from the subjects through an electronic medical system. Demographics include age, sex, duration of diabetes mellitus, smoking and alcohol consumption. Anthropometric measurements include body mass index (BMI), abdominal circumference (AC), systolic blood pressure (SBP), and diastolic blood pressure (DBP). Clinical parameters included high-density lipoprotein cholesterol (HDL-C), fasting blood glucose (FBG), serum creatinine (SCR) and glycated hemoglobin (HbA1c). Disease conditions include hypertension, diabetic complications and stroke. Medication use includes antidiabetic drugs, lipid-lowering drugs, laboratory parameters and antihypertensive drugs.

## Study variables

BMI is calculated by dividing body weight (kg) by the square of height (m). The World Health Organization (WHO) classification criteria for BMI in Asia are: BMI <18.5kg/m2 is considered underweight, normal weight is 18.5-24.0 kg/m2, overweight is 24.0-28.0 kg/m2, and obesity is >29.0 kg/m2.[13] According to the recommendations of the American Diabetes Association,[14] HbA1c ⩾7% is defined as hyperglycemia, and HDL-C ⩽1 mmol/L in men and HDL-C ⩽1.3 mmol/L in women were defined as dyslipidemia, all of which indicated that treatment goals were not met. The formula for calculating glomerular filtration rate (eGFR) is:[15]

$$\text{Female: GFR} = \begin{cases} 144 \times (SCR/0.7)^{-0.329} \times 0.993^{age}, SCR \leq 0.7 mg/dl \\ 144 \times (SCR/0.7)^{-1.209} \times 0.993^{age}, SCR > 0.7 mg/dl \end{cases}$$

$$\text{Male: GFR} = \begin{cases} 141 \times (SCR/0.9)^{-0.411} \times 0.993^{age}, SCR \leq 0.9 mg/dl \\ 141 \times (SCR/0.9)^{-1.209} \times 0.993^{age}, SCR > 0.9 mg/dl \end{cases}$$

The overall statistical analysis process of this paper is shown in Figure 1. A preprocessing method is mainly included and investigated. The preprocessing process includes the elimination of missing values as well as feature selection, the optimization of hyperparameters using grid search, and the evaluation and analysis of classifiers. In addition, a 10-fold cross-validation is used to avoid the effect of dividing the training set and the test set differently.

## *Statistical Analysis*

### Data preprocessing

In the prediction model, whether DN occurs or not is defined as a binary variable. Illness is denoted as 1; absence of illness is denoted as 0. The features with more than 50% missing values were excluded, and then the samples with missing values were removed from the analysis (see Figure 2). In addition, in this paper, the features are divided into continuous and categorical variables for data preprocessing. If the features are continuous, they are normalized using the Z-Score method. If the features are categorical and there is no size significance between the fetched values, the fetched values of the discrete features are extended to the Euclidean space using the unique hot coding (One-Hot).

### Feature Selection

Feature selection was performed by using least absolute shrinkage and selection operator (LASSO) regression. The Lasso regression model improves the prediction performance by adjusting the hyperparameter λ to compress the regression coefficients to zero and selecting the feature set that performs best in DN prediction. To determine the best λ value, λ was selected by minimum mean error using 10-fold cross-validation.

### Model Training and Validation

In this paper, the 10-fold cross-validation method is used to divide the training and testing sets, i.e., in each cycle, 9 subsets are used as the training set and 1 subset is used as the testing set. The model is optimized by using grid search.   DN prediction models were using 10-fold cross-validation as a model evaluation strategy, and four classification algorithms,

Extreme Gradient Boosting (XGB), random forest (RF), decision tree (DT), and logistic regression, respectively, mainly for predicting the risk of diabetic nephropathy in individuals. The above models are evaluated based on their generalization ability and practicality. The generalization ability of the model is examined by the receiver operating characteristic (ROC) curve, and the area under the curve (AUC) values of the model, and the clinical utility of the model was examined by using the decision curve and calibration curve.

# Analysis of results

## *Pre-processing results*

Through the above missing value processing (see Section 2.2.1), the final size of the dataset was obtained as 562 × 119 (number of samples × number of features), which is sufficient sample size to meet the statistical requirements and ensure the reliability of the study results.[16,17]

The clinical characteristics of the participants according to DN as a column stratified variable are shown in Table 1. The presence or absence of DN is statistically significant with HDL, Apo AI, C4DC, C5DC, HbA1c, and Hypertension ( $p < 0.05$ ). Compared with non-diabetic renal disease (NDRD), patients with DN tends to be without hypertension, with hyperglycemia, as well as have higher levels of HDL, Apo AI, C5DC, and lower levels of C4DC.

Feature screening. Based on the 'glmnet' package implementation in R language, the best performing features were screened from 70 clinical information and 49 metabolic indicators to reduce the dimensionality; therefore, the predictive performance of the classifier was significantly improved. After Lasso regression screening (see Figure 3), the best feature set was obtained including clinical information: diabetes duration, AC, SBP, hemoglobin concentration (HB), erythrocyte pressure volume (PCV), globulin (GLB), alkaline phosphatase (ALP), blood uric acid (UA), urinary microalbumin (MAU), cholesterol (CHOL), HDL, apolipoprotein AI (Ap0AI), and Apo B (AI0B), insulin (INS), FBG, glutamic acid decarboxylase antibody (GADA), insulin sample growth factor-1 (IGF-1), free triiodothyronine (FT3), thyroid stimulating hormone (TSH), eGFR, HbA1c, hypertension (high blood pressure was recorded as 1 and vice versa, as 0) and Thiazolidinediones (TZDs), and Glinides (Glinides), lipid-lowering drugs, dipeptidyl peptidase-4 (DPP-4), glucagon-like polypeptide (GLP_1), and sodium-glucose co-transport protein 2 inhibitor (SGLT-2); amino acids including cysteine (Cys), methionine (Met), serine (Ser), and tyrosine (Tyr); acylcarnitine including acetyl carnitine (C2), succinyl carnitine (C4DC), glutaryl carnitine (C5DC), and tetracosanoic carnitine (C24).

## *Hyperparameter optimization results*

In this study, based on GridSearchCV in sklearn, for each combination in the hyperparameter combination list, four different machine learning models are instantiated, 10-fold cross-validation is done, and the parameter combination with the highest average score is returned using 'roc_auc' as the scoring criterion, as shown in Table 1.

## *Classifier results*

Based on the pre-processed Dalian dataset, the four classifiers of XGB, RF, DT and logistic regression were used to classify diabetic nephropathy, which showed that the XGB model was significantly better than the RF, logistic regression and DT models. The AUC value of the DT model was greater than 0.8, but the false positive rate was higher than the other three models, so it was not recommended.

The decision curve provides an adequate representation of the clinical utility of a model, i.e., at a certain threshold probability, the net benefit of the model is higher than the two special cases of no intervention for anyone and intervention for everyone at the same time, indicating that the model has practical value. As shown in Figure 5, all models were valid between the thresholds of 28%-81%, and between the thresholds of 11%-86%, the net benefit of the XGB model outperformed the other three models.

A new sample data set was obtained by bootstrap method using Python 3.10 by sampling 10,000 times independently to plot the calibration curve of XGB model. As shown in Figure 6, after the XGB model was calibrated, the curve gradually approached the diagonal line, indicating that the screening is close to the real situation and has practical value.

## *Model interpretation*

The effect of features on screening scores is measured by SHAP, which evaluates the importance of each feature using a game-theoretic approach based on the test set.[18] When the Shapley value of each feature is positive, it indicates an increased risk of DN; conversely, it indicates a decreased risk of DN. The scattering colors in the figure indicate the magnitude of the feature values, with red being larger and blue being smaller. As shown in Figure 7, MAU, diabetes duration, PVC, FPG and eGFR contributed more to the model; in the metabolite group, C2, C5DC, Tyr, Ser and Met contributed more to the model.

When the duration of diabetes is greater than or equal to 15, the threshold value of Tyr that best describes the difference in outcomes is 45, at which point the higher the Tyr value, the lower the risk of DN (as shown in Figure 8(c)). In addition, patients with longer diabetes duration and lower C5DC values had a lower risk of disease compared to those with higher C5DC values; patients with longer diabetes duration and lower Tyr values had a higher risk of disease compared to those with higher Tyr values; or patients with lower C24 values and compared to those with higher Tyr values and longer diabetes duration; C24 vs. C5DC reasoning was the same.

## Discussion

This study focuses on the metabolites, where C2, C5DC, Tyr, Ser, Met, C24, C4DC, and Cys have a strong effect on DN and can be used as new biomarkers for DN.

Aromatic amino acids are a group of α-amino acids that contain an aromatic ring, including phenylalanine, tyrosine and tryptophan. Phenylalanine is oxidized to tyrosine by phenylalanine hydroxylase and then involved in glucose metabolism.[24] In a prospective study, lower plasma tyrosine levels in diabetic patients were associated with an increased risk of microvascular disease.[25] A previous study confirmed the association between low tyrosine concentrations and diabetic nephropathy.[26]

Methionine is an essential sulfur-containing amino acid that is required for normal growth and development of the body and is also associated with %FM. It is a precursor of succinyl CoA, homocysteine, creatine and carnitine, which the organism generally obtains from food or gastrointestinal microorganisms. Methionine plays a crucial role in the immune system because its catabolism leads to increased production of glutathione, taurine and other serum metabolites.[27] Methionine and other methyl donors improve glucose tolerance and insulin sensitivity in the offspring of high-fat diet mice.[28] Experiments in rats have demonstrated that methionine ameliorates alterations in key one-carbon serum metabolites and T2D-induced disturbances in glucose and lipid metabolism in T2D rats.[29] And there is growing evidence that methionine activates AMPK and SIPT1 by a mechanism similar to that of metformin.[30] Given that diabetic nephropathy is one of the microvascular complications of type 2 diabetes, it is reasonable to speculate that methionine disorders are negatively associated with type 2 diabetes complicated by diabetic nephropathy.

Diabetes mellitus as a metabolic dysfunctional disease damages several organs and systems, including the liver, kidneys and peripheral nerves. Although essential amino acids are important for maintaining normal physiological activities of the body, abnormal metabolism of nonessential amino acids is also associated with the pathogenesis of diabete.[31,32] Serine, a non-essential amino acid, levels have been found to be consistently reduced in patients with metabolic syndrome.[33] In a prospective study, elevated serum glycine levels were found to be associated with a reduced risk of developing type 2 diabetes.[34] Glycine being a precursor substance of serine,[35] there is even more reason to speculate about the importance of serine in the microvascular complications of type 2 diabetes.

Numerous studies have found that homocysteine, a precursor substance of cysteine, is considered a biomarker for microvascular diseases including diabetic neuropathy, retinopathy and nephropathy-like diseases.[36] Epidemiological studies have shown a U-shaped relationship between cardiovascular disease and cysteine after adjusting for other risk factors and homocysteine.[37] In this study, screening metabolic indicators associated with diabetic nephropathy by the LASSO model revealed a positive association between cysteine and diabetic nephropathy; the fact that no risk trend relationship was observed in the first half of the U-shaped curve may be due to the fact that this study was conducted based on type 2 diabetic patients, who have much higher levels of oxidative stress and reactive oxygen species than normal subjects.

Acylcarnitine is known to play a key role in the β-oxidation of long-chain fatty acids through the inner mitochondrial membrane. Comparing cases of obesity, insulin resistance, metabolic syndrome and diabetes with relevant controls revealed that acylcarnitine was characterized differently between groups. A 6-year prospective study of 2103 community-dwelling individuals aged 50-70 years in Beijing and Shanghai, China, with type 2 diabetes as the observed outcome found higher plasma concentrations of short-, medium-, and long-chain acylcarnitines at baseline, but only long-chain acylcarnitines were significantly associated with the risk of type 2 diabetes.[38] A previous study found that elevated levels of short- and medium-chain acylcarnitines in blood were associated with the risk of developing Cardiovascular disease in T2DM.[39] A study on diabetic peripheral neuropathy (DPN) claimed that C4DC and C24 concentrations in non-DPN plasma were significantly higher than in DPN patients, and that factors containing C2, C3, C4, and C5 short-chain acyl carnitines were positively associated with the risk of DPN in T2DM.[40] C2 is derived from carbohydrate catabolism and acetyl-CoA, the end product of β-oxidation.[41] It was also found that C2 may be a biomarker of combined sugar and lipid toxicity. And animal experiments also showed that plasma C2 levels were elevated in T2DM rats.[42]

Proteinuria and eGFR loss are both nonspecific markers of DN, but have limitations as prognostic tools.[19] This is because a high percentage of T2DM patients in renal biopsy studies do not have DN and suffer from other renal diseases.[20] Therefore, it is important to identify new prognostic markers for DN based on serum metabolites in this paper. However, due to the limitation of data, this paper is limited to the dichotomous problem, and the multiclassification model for DN grade can be further investigated in the future.

## Conclusion

This paper constructs a XGB model to screen for DN, whose predictive performance is better than those in previous studies [21-23] with 0.93, 0.79, 0.90. Lasso plays a key role in ensuring the accuracy and stability of the predictive model, which improves the quality of the dataset. C2, C5DC, Tyr, Ser, Met, C24, C4DC, and Cys are shown to be highly correlated with DN risk.

This paper introduces serum metabolites as new DN markers, constructs several machine learning models to screen for DN, compares their screening abilities, and analyzes the impact of each important feature on DN. The results show that the XGB model has the best screening effect, and Lasso model plays a key role in ensuring the accuracy and stability of the screening model, which improves the quality of the dataset. In addition, compared with previous studies [21-23], our model has better result.

# Acknowledgments

## Funding

This work was supported by the National Key Research and Development Program of China (2021YFA1301202), the National Natural Science Foundation of China (82273676), the Liaoning Province Scientific and Technological Project (2021JH2/10300039), and the Science & Technology Development Fund of Tianjin Education Commission for Higher Education [2022KJ204].

## Medical Writing, Editorial, and Other Assistance

The authors thank all the doctors, nurses, and research staff at the SAHDMU in Dalian, for their participation in this study. Editorial assistance in the preparation of this article was provided by Dr. Yang Li of Tianjin Medical University.

## Author Contributions

All authors contributed to the study conception and design. Material preparation, data collection and analysis were performed by Jing-Mei Yin, Yang Li and Guo-Wei Zong. The first draft of the manuscript was written by Jing-Mei Yina and all authors commended on precious versions of the manuscript. All authors read and approved the final manuscript.

## Conflict of Interest

The authors declare that there is no conflict of interest regarding the publication of this paper.

## Data Availability

The datasets generated during and analyzed during the current study are available from the corresponding author on reasonable request.

**Table 1** Baseline characteristics of the study population

|  | NDRD | DN | P value |
|---|---|---|---|
| Duration of T2D | 8.99 ± 54.68 | 14.26 ± 84.77 | 0.10 |
| AC | 91.98 ± 93.95 | 94.97 ± 99.70 | 0.96 |
| SBP | 144.31 ± 392.02 | 153.09 ± 482.23 | 0.51 |
| Hb | 143.08 ± 217.92 | 135.98 ± 405.09 | 0.24 |
| PCV | 42.46 ± 29.35 | 40.16 ± 37.49 | 0.26 |
| GLB | 27.35 ± 17.11 | 28.93 ± 19.20 | 0.42 |
| ALP | 74.01 ± 778.13 | 72.67 ± 490.06 | 0.56 |
| UA | 327.36 ± 7740.74 | 363.47 ± 11384.12 | 0.26 |
| MAU | 31.65 ± 8476.00 | 245.03 ± 245357.87 | 0.33 |
| CHOL | 5.01 ± 1.24 | 5.09 ± 2.48 | 0.33 |
| HDL | 2.57 ± 0.63 | 2.48 ± 0.76 | 0.02 |
| ApoAI | 1.41 ± 0.05 | 1.38 ± 0.05 | < 0.01 |
| ApoB | 1.00 ± 0.18 | 1.01 ± 0.12 | 0.29 |
| INS | 14.52 ± 174.52 | 21.75 ± 446.94 | 0.36 |
| FBG | 8.61 ± 8.90 | 9.98 ± 14.94 | 0.77 |
| GADA | 6.51 ± 174.38 | 7.46 ± 251.27 | 0.32 |
| IGF-1 | 162.61 ± 3073.97 | 151.01 ± 3473.32 | 0.50 |
| FT3 | 4.84 ± 0.82 | 4.59 ± 0.95 | < 0.01 |
| TSH | 2.15 ± 4.69 | 3.29 ± 93.22 | 0.77 |
| Cys | 1.42 ± 0.70 | 1.46 ± 0.82 | 0.88 |
| Met | 15.10 ± 17.93 | 14.15 ± 14.84 | 0.39 |
| Ser | 45.14 ± 135.16 | 44.89 ± 118.29 | 0.85 |
| Tyr | 51.36 ± 255.07 | 46.92 ± 252.05 | 0.27 |
| C2 | 11.60 ± 17.59 | 12.45 ± 19.78 | 0.93 |
| C4DC | 0.38 ± 0.04 | 0.37 ± 0.03 | < 0.01 |

Table 1 (continued)

| | NDRD | DN | P value |
|---|---|---|---|
| C5DC | 0.06 ± 0.001 | 0.07 ± 0.001 | < 0.01 |
| C24 | 0.04 ± 0.0003 | 0.04 ± 0.0004 | < 0.01 |
| eGFR | 95.81 ± 504.53 | 84.97 ± 976.44 | 0.48 |
| HbA1c (%) | | | |
| 0-7 | 88 (31.10) | 44 (15.77) | < 0.01 |
| ⩾7 | 195 (68.9) | 235 (84.23) | |
| AGI (%) | | | |
| Yes | 188 (66.43) | 147 (52.69) | < 0.01 |
| No | 95 (33.57) | 132 (47.31) | |
| TZDs (%) | | | |
| Yes | 272 (96.11) | 263 (94.27) | 0.41 |
| No | 11 (3.89) | 16 (5.73) | |
| Glinides (%) | | | |
| Yes | 260 (91.87) | 271 (97.13) | 0.01 |
| No | 23 (8.13) | 8 (2.87) | |
| Dpp-4 (%) | | | |
| Yes | 266 (93.99) | 257 (92.11) | 0.48 |
| No | 17 (6.01) | 22 (7.89) | |
| GLP-1 (%) | | | |
| Yes | 279 (98.59) | 269 (96.42) | 0.17 |
| No | 4 (1.41) | 10 (3.58) | |
| SGLT-2 (%) | | | |
| Yes | 282 (99.65) | 275 (98.57) | 0.36 |
| No | 1 (0.35) | 4 (1.43) | |

Table 1 (continued)

| | NDRD | DN | P value |
|---|---|---|---|
| Hypertension (%) | | | |
| Yes | 155 (54.77) | 102 (36.56) | < 0.01 |
| No | 128 (45.23) | 177 (63.44) | |
| Lipid-lowering drug (%) | | | |
| Yes | 169 (92.35) | 133 (74.30) | < 0.01 |
| No | 14 (7.65) | 46 (25.7) | |
| Drink (%) | | | |
| Yes | 275 (97.17) | 268 (96.06) | 0.62 |

|  | No | 8 (2.83) | 11 (3.94) |
| --- | --- | --- | --- |

Mean ± SD for continuous variables: the p-value was calculated by a t test. % for categorical variables: the p-value was calculated by a weighted chi-square test. MAU, urinary microalbumin; Duration of T2D, duration of type 2 diabetes mellitus; PCV, erythrocyte pressure volume; SBP, systolic blood pressure; FBG, fasting blood glucose; eGFR, glomerular filtration rate; IGF-1, Insulin Sample Growth Factor-1; INS, Insulin; FT3, Free Triiodothyronine; GLB, Globulin; Hb, Hemoglobin Concentration; CHOL, Cholesterol; ALP, Alkaline Phosphatase; UA, Blood Uric Acid; HbA1c, Glycosylated Hemoglobin; HDL, High Density Lipoprotein; Tyr, Chitosan; Ser, Serine; AC, Abdominal Circumference; Met, Methionine; Ap0AI, Apolipoprotein A1; AP0B, Apolipoprotein B; TSH, Thyroid Stimulating Hormone; AGI, Alpha-Glucosidase Inhibitor; GADA, Glutamic Acid Decarboxylase Antibody; Cys, Cysteine; drink: alcohol consumption or not; Dpp-4, dipeptidyl peptidase-4; TZDs: thiazolidinediones; GLP-1, glucagon-like peptide; SGLT-2, sodium-glucose cotransporter protein 2 inhibitor; C2, acetylcarnitine; C4DC, succinylcarnitine; C5DC, glutarylcarnitine; C24, tetracosanoic carnitine.

**Table 2** Highest AUC scores achieved by hyperparameter tuning of four machine learning models

| Classifiers | Adjusted hyperparameters | Accuracy | AUC | P value |
| --- | --- | --- | --- | --- |
| XGB | learning_rate=0.001,n_estimators=1500, max_depth=3,subsample=0.6,colsample_bytree=0.6,gamma=0.3,objective='bina | 0.875 | 0.966 | .000 |
| | ry:logistic',scale_pos_weight=1,reg_lambda=0.1 | | | |
| RF | max_depth=9，min_samples_split=5，min_samples_leaf=，n_estimators=170 | 0.875 | 0.937 | .000 |
| DT | max_depth=3,min_samples_split=3，in_samples_leaf=1 | 0.8929 | 0.812 | .000 |
| Logistic | penalty='l1'，C=0.8，tol=0.0001 | 0.8036 | 0.845 | .000 |

XGB, eXtreme Gradient Boosting; RF, Random Forest; DT, Decision Tree. C indicates the inverse of the regularization factor; tol indicates that the calculation stops when the solution reaches 0.0001 and the optimal solution is considered to have been found; P values are the Delong's test for the area under the receiver operating characteristic curve, and <0.05 indicates that the analysis method is statistically significant.

**Notes:**

**Abbreviations:** AC, abdominal circumference; ALP, Alkaline Phosphatase; AGI, Alpha-Glucosidase Inhibitor; Ap0AI, Apolipoprotein A1; Ap0B, Apolipoprotein B; AUC, area under the curve; BMI, body mass index; C2, acetylcarnitine; C4DC, succinylcarnitine; C5DC, glutarylcarnitine; C24, tetracosanoic carnitine; CHOL, Cholesterol; Cys, Cysteine; DN, diabetic nephropathy; DT, Decision Tree; DBP, diastolic blood pressure; DCA, Decision Curve; Duration of T2D, duration of type 2 diabetes mellitus; drink: alcohol consumption or not; Dpp-4, dipeptidyl peptidase-4; eGFR, glomerular filtration rate; FT3, Free Triiodothyronine; FBG, fasting blood glucose; FBG, fasting blood glucose; GLB, Globulin; GADA, Glutamic Acid Decarboxylase Antibody; GLP-1, glucagon-like peptide; HDL-C, high-density lipoprotein cholesterol; HbA1c, glycated hemoglobin; HB, Hemoglobin Concentration; HbA1c, Glycosylated Hemoglobin; IGF-1, Insulin Sample Growth Factor-1; INS, Insulin; LASSO, Least absolute shrinkage and selection operator; Met, Methionine; MAU, urinary microalbumin; NNs, Artificial Neural Networks; PCV, erythrocyte pressure volume; ROC, receiver operating characteristic; RF, Random Forest; SBP, systolic blood pressure; SCR, serum creatinine; SGLT-2, sodium-glucose cotransporter protein 2 inhibitor; Ser, Serine; SVM, Support Vector Machine; Tyr, Chitosan; TSH, Thyroid Stimulating Hormone; TZDs: thiazolidinediones; UA, blood uric acid; XGB, eXtreme Gradient Boosting.

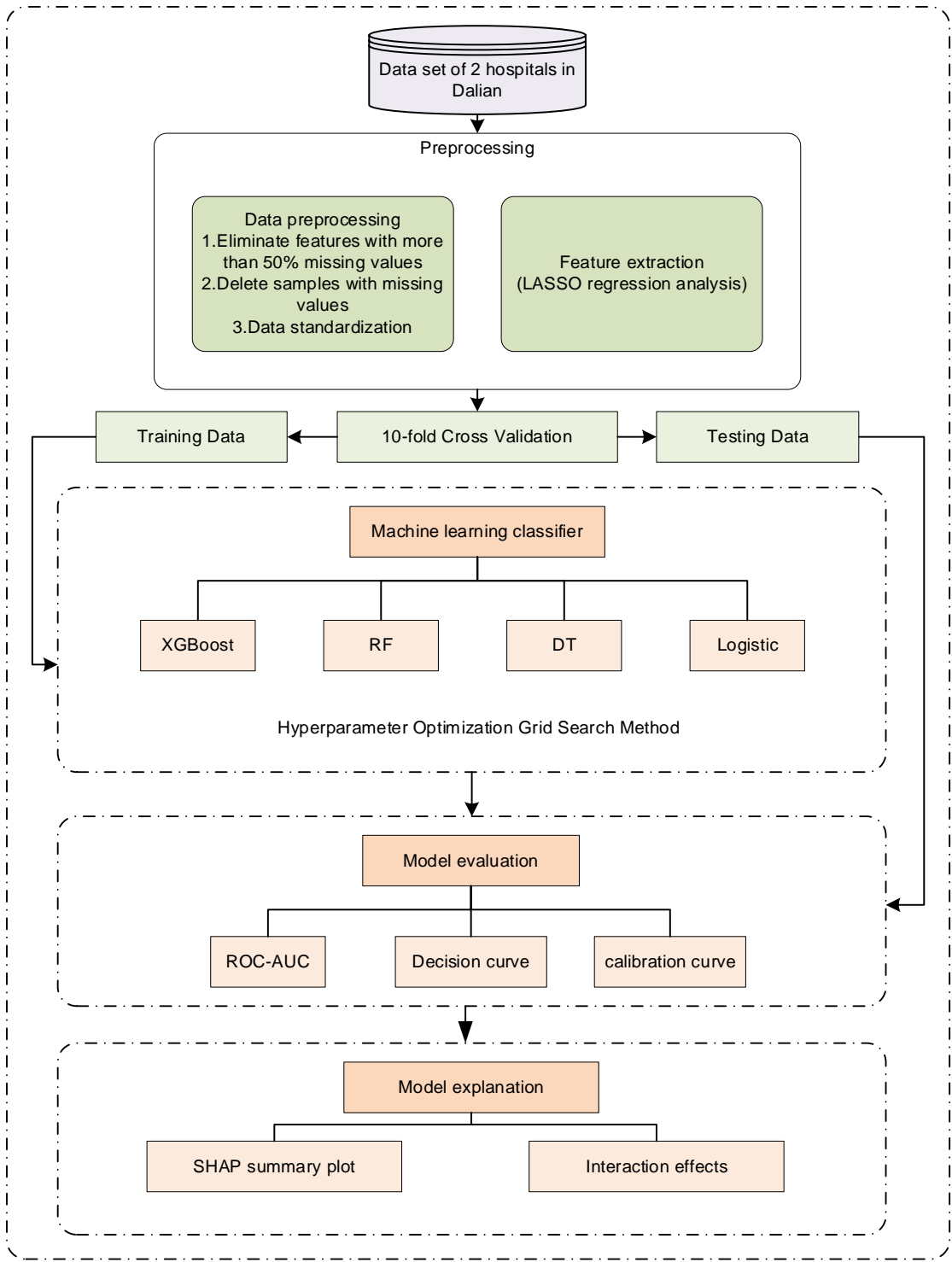

**Figure 1** DN statistical analysis workflow diagram

DN statistical analysis workflow diagram contains four machine learning classifiers, preprocessing steps, optimization of hyperparameters of classifiers by grid search, and model evaluation methods. Feature filtering was performed using R V4.2.2. Data preprocessing and modeling, evaluation and interpretation of machine learning models were performed with Python V3.10.

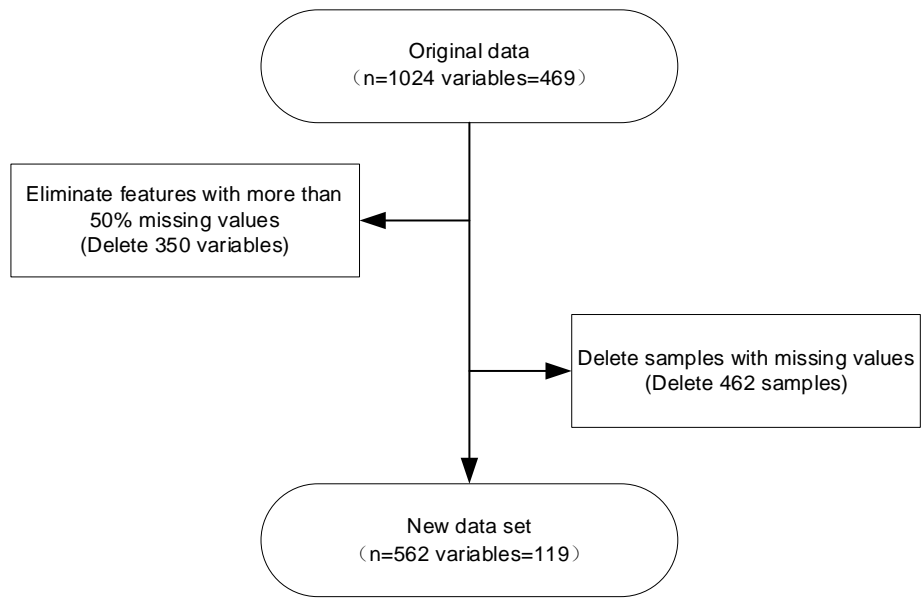

**Figure 2** Data pre-processing process.

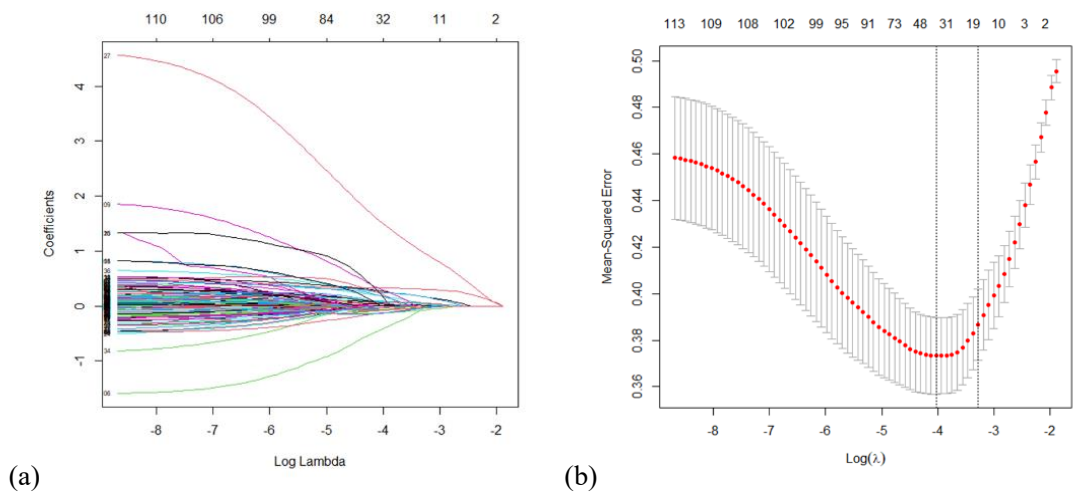

(a)　　　　　　　　　　　　　　　　(b)

**Figure. 3** (a) Lasso coefficient profiles of 119 features, (b) the value of λ with the smallest mean error is selected by 10-fold cross-validation

(a) each line represents a feature, and each estimated parameter decreases as λ increases until it compresses to 0. (b) The relationship between the mean square error and log(λ) is plotted. Vertical dashed lines are plotted at the best value using the minimum criterion and the 1SE principle. Based on 10-fold cross-validation, the λ value of 0.01783 was selected and the optimal number of features was obtained as 38.

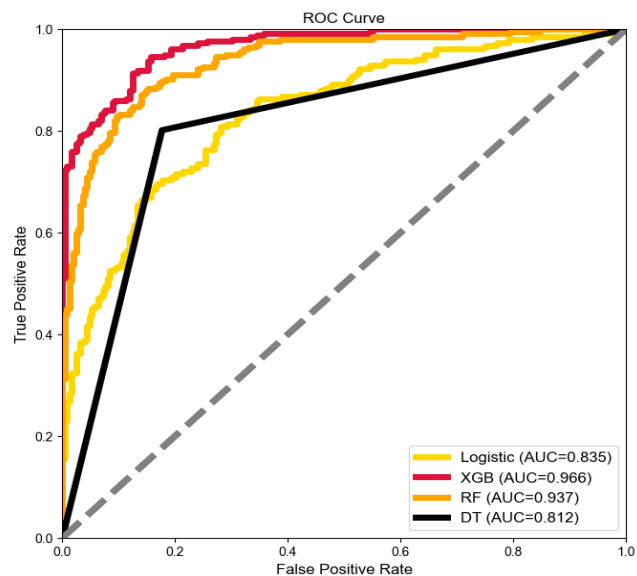

**Figure 4** ROC curves of the four models based on the test set data

XGB, eXtreme Gradient Boosting; RF, Random Forest; DT, Decision Tree. The red line indicates the XGB model based on the Lasso selection feature (AUC=0.966), the orange line indicates the RF model based on the Lasso selection feature (AUC=0.937), the yellow line indicates the Logistic model based on the Lasso selection feature (AUC=0.845), and the black line indicates the DT model based on the Lasso selection feature (AUC= 0.812).

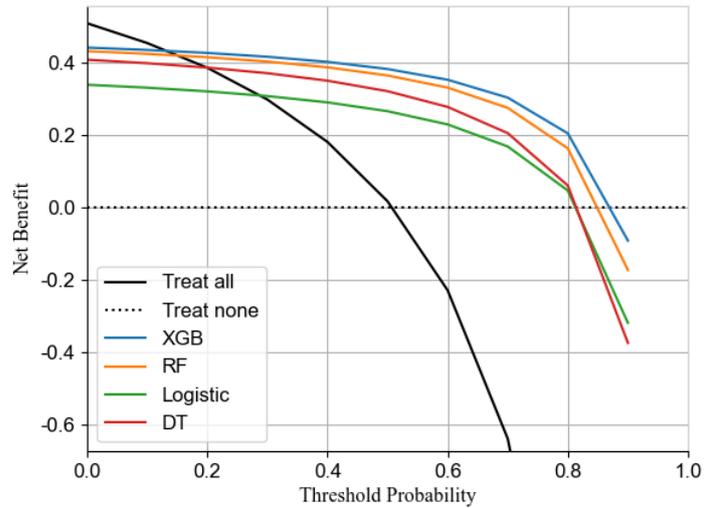

**Figure 5** Clinical utility of the 4 models

XGB, eXtreme Gradient Boosting; RF, Random Forest; DT, Decision Tree. the dashed line indicates the net benefit when intervening on no one and the black curve indicates the net benefit when intervening on everyone. The blue line indicates the net benefit in the case of the XGB model, the orange line indicates the net benefit in the case of the RF model, the green line indicates the net benefit in the case of the Logistic model, and the red line indicates the net benefit in the case of the DT model.

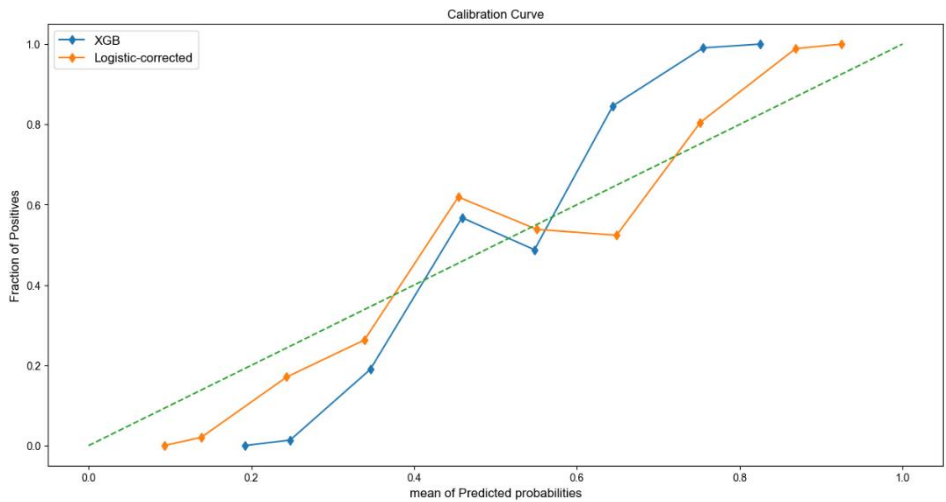

**Figure 6** Calibration curves of the XGB model

XGB, eXtreme Gradient Boosting; Logistic-corrected, corrected XGB model. The blue line indicates before calibration of the XGB model based on the Lasso selection feature, the orange line indicates after calibration of the XGB model based on the Lasso selection feature, and the dashed line indicates full calibration.

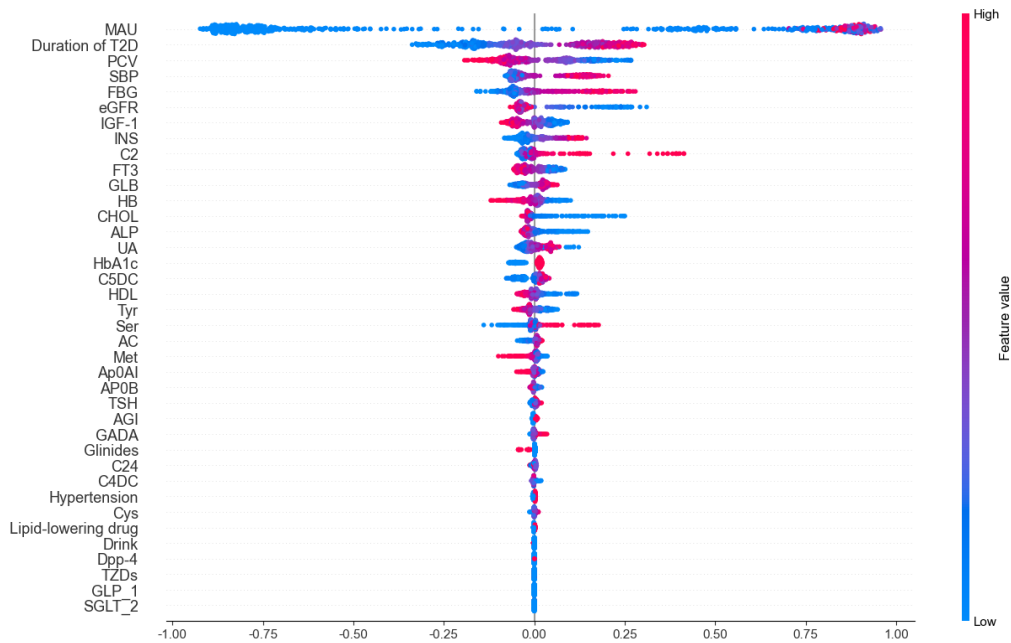

**Figure 7** SHAP summary plot for the XGB model based on Lasso selection of features

XGB, eXtreme Gradient Boosting; RF, Random Forest; DT, Decision Tree. Each point on the summary plot is the Shapley value of the feature and the instance. The position on the y-axis is determined by the feature, and the x-axis is determined by the Shapley value determination. Colors indicate feature values from low to high. The features are arranged according to their importance. SHAP, Shapley Additive exPlanations.

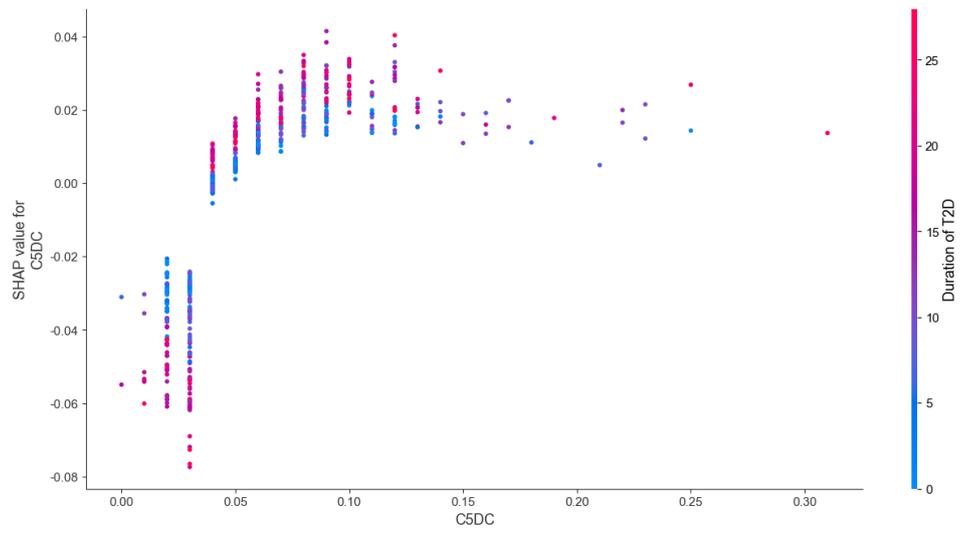

(a)

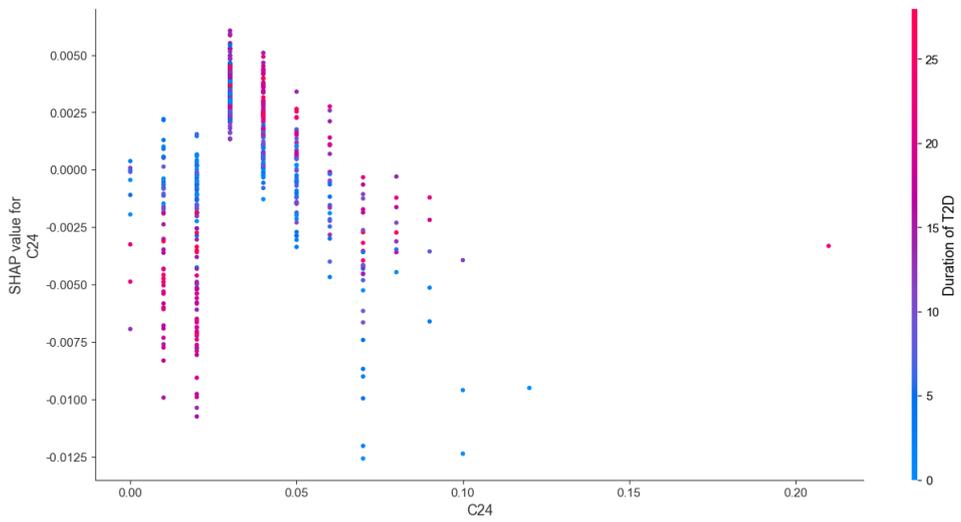

(b)

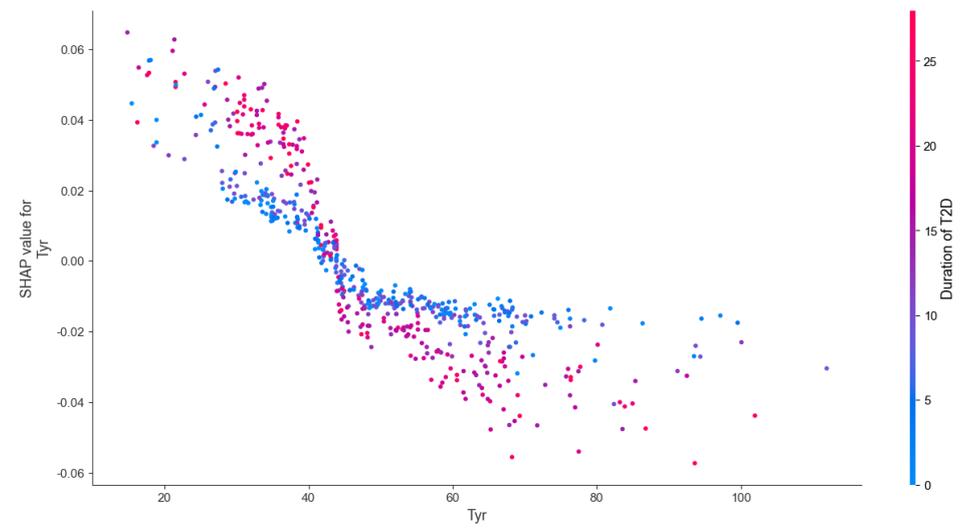

(c)

**Figure 8** SHAP plot showing the nonlinear interaction

It shows the nonlinear interaction between diabetes duration and serum metabolites, including C5DC (a), C24 (b), and Tyr (c). X-axis indicates the value of the feature, and y-axis indicates the Shapley value of the feature. Red indicates a larger right-hand feature, and blue indicates a smaller right-hand feature. The Shapley value indicates the effect of the feature on the model.